\pdfoutput=1

\documentclass[11pt]{article}

\usepackage[]{EMNLP2023}

\usepackage{times}
\usepackage{latexsym}

\usepackage[T1]{fontenc}
\usepackage[utf8]{inputenc}

\usepackage{microtype}

\usepackage{color}
\usepackage{amsmath}
\usepackage{graphicx}

\usepackage{arydshln} 
\usepackage{multirow}

\definecolor{c1}{HTML}{95bddc}
\definecolor{c2}{HTML}{c2d1e5}
\definecolor{c3}{HTML}{fe793d}
\definecolor{c4}{HTML}{fb4c1f}
\definecolor{c5}{HTML}{b71a3b}
\definecolor{c6}{HTML}{7e0f12}
\definecolor{c7}{HTML}{E85642}
\definecolor{c8}{HTML}{C00000}
\definecolor{c9}{HTML}{ff2d51}

\title{Exploiting Emotion-Semantic Correlations for \\Empathetic Response Generation}

\author{
Zhou Yang$^{1,2}$, Zhaochun Ren$^{3}$, Yufeng Wang$^{1,2}$, Xiaofei Zhu$^{4}$, Zhihao Chen$^{1,2}$, \\
\textbf{Tiecheng Cai}$^{1,2}$, \textbf{Yunbing Wu}$^{1,2}$, \textbf{Yisong Su}$^{1,2}$, \textbf{Sibo Ju}$^{1,2}$, 
\textbf{Xiangwen Liao}$^{1,2}$\thanks{\hspace{1mm} Corresponding author.}\\
\small $^1$College of Computer and Data Science, Fuzhou University; 
$^2$Digital Fujian Institute of Financial Big Data,
 Fuzhou, China \\ 
\small $^3$Leiden University, Leiden, The Netherlands \\ 
\small $^4$College of Computer Science and Technology, Chongqing University of Technology, Chongqing, China\\ 
\small  \texttt{\{200310007, 211027083, n180320046, 210310002, 221010003, 221010003\}@fzu.edu.cn} \\
\small  \texttt{z.ren@liacs.leidenuniv.nl} \hspace{0.1cm} \texttt{zxf@cqut.edu.cn} \hspace{0.1cm} \texttt{\{wyb5820, liaoxw\}@fzu.edu.cn}\\
}

\begin{document}
\maketitle
\begin{abstract}
Empathetic response generation aims to generate empathetic responses by understanding the speaker's emotional feelings from the language of dialogue.
Recent methods capture emotional words in the language of communicators and construct them as static vectors to perceive nuanced emotions. 
However, linguistic research has shown that emotional words in language are dynamic and have correlations with other grammar semantic roles, i.e., words with semantic meanings, in grammar. Previous methods overlook these two characteristics, which easily lead to misunderstandings of emotions and neglect of key semantics.

To address this issue, we propose a dynamical Emotion-Semantic Correlation Model (ESCM) for empathetic dialogue generation tasks. ESCM constructs dynamic emotion-semantic vectors through the interaction of context and emotions. We introduce dependency trees to reflect the correlations between emotions and semantics. Based on dynamic emotion-semantic vectors and dependency trees, we propose a dynamic correlation graph convolutional network to guide the model in learning context meanings in dialogue and generating empathetic responses. Experimental results on the EMPATHETIC-DIALOGUES dataset show that ESCM understands semantics and emotions more accurately and expresses fluent and informative empathetic responses. Our analysis results also indicate that the correlations between emotions and semantics are frequently used in dialogues, which is of great significance for empathetic perception and expression.\footnote{Our code is available at \url{https://github.com/zhouzhouyang520/EmpatheticDialogueGeneration_ESCM}}
\end{abstract}
\section{Introduction}
Attracting an increasing amount of attention, empathetic response generation aims to generate empathetic responses by perceiving the speaker's emotional feelings~\cite{rashkin2018towards,zhong2021care,zhong2020towards,liang2021infusing,2021comae,liu2021towards,wang2021cass}.

Early methods perceive the speaker's feelings by understanding the holistic semantics and emotions expressed in the language of the context
~\cite{rashkin2018towards, lin2019moel, majumder2020mime}.
These methods are prone to generate trivial and uninformed responses, ascribed to the neglect of nuances of human emotion in dialogues~\cite{li2019empdg}.
To address this issue, recent methods detect emotional words in the language of the communicators and build them as static vectors to perceive subtle emotions
~\cite{li2019empdg, CEM2021, Kim2021empathy, gao2021improving, li-etal-2022-kemp, kim2022emp}.

\begin{figure}
\centering
\includegraphics[width=82mm]{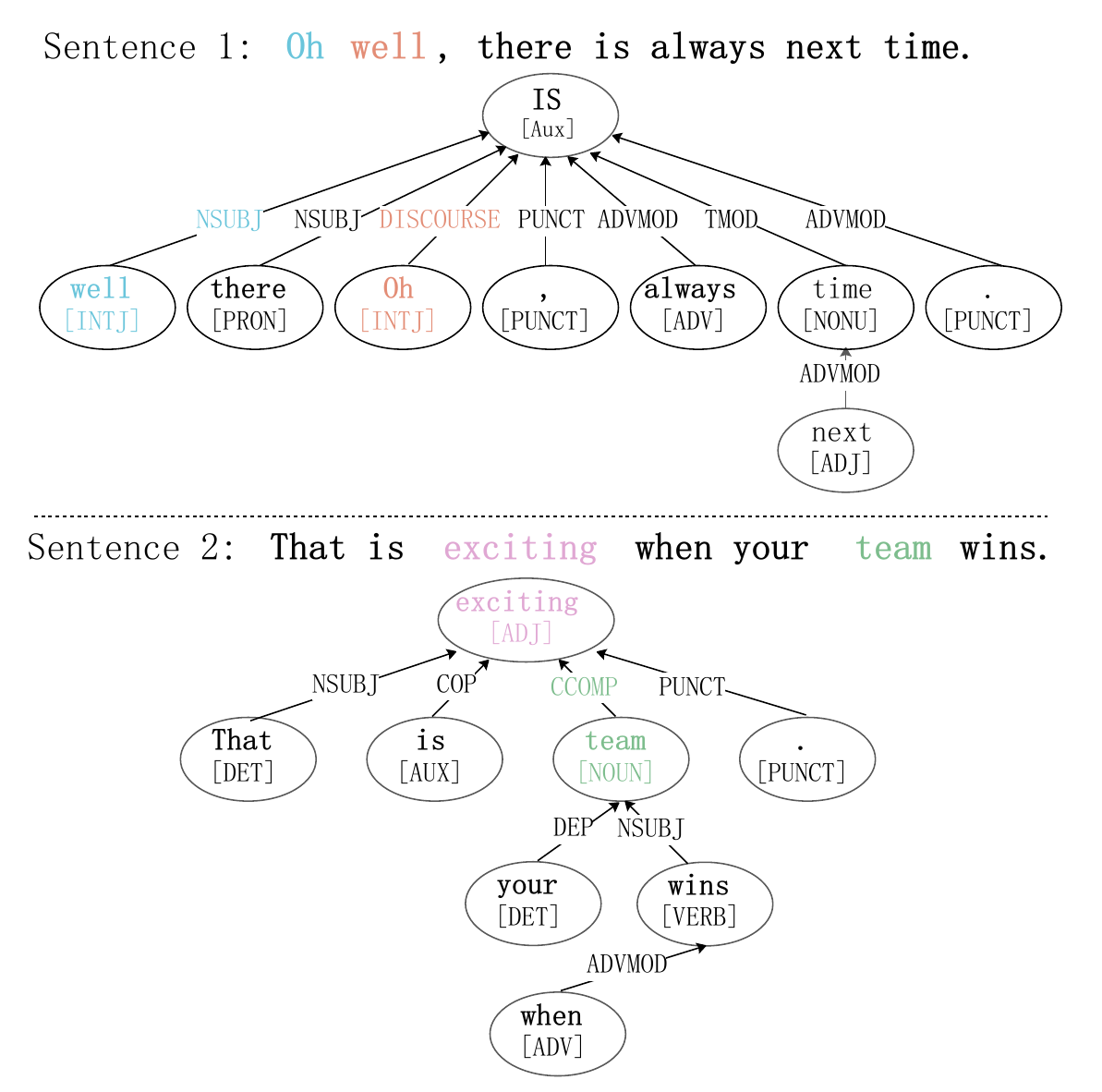}
\caption{\label{fig 1}
Examples from the EMPATHETIC-DIALOGUES dataset.
Sentence 1 shows the variability of emotional words. Sentence 2 shows the correlations of emotional words with semantic roles.
}
\end{figure}

However, according to linguistic research~\cite{foolen2012moving, dirven1997emotions, osmond1997prepositions, radden1998conceptualisation}, focusing on emotional words while ignoring their characteristics in the expression process leads to emotional misunderstandings and the neglect of words with important semantic information.
As an important theory of emotional expression in linguistics, the conceptualization of emotions~\cite{foolen2012moving} suggests that emotional words have two important characteristics in the expression process: variability and relevance. 
Variability is that the affection of emotional words changes dynamically during the expression process.
For example, ``well'' generally carries a positive meaning, but in sentence 1 (shown in Figure \ref{fig 1}), it is used as an interjection to express a neutral emotion. 
Using static vectors (such as Embedding~\cite{pennington2014glove,mikolov2013word2vec} or VAD~\cite{mohammad2018obtaining}) to represent this dynamic emotion, previous methods are prone to misunderstand this sentence as positive.

Relevance refers to the grammatical correlations between emotional words and words carrying semantic meaning, which plays an important role in understanding emotions and semantics.
For example, as shown in Figure \ref{fig 1}, sentence 2 expresses the ``exciting'' emotion due to the victory of ``team''. ``Team'' is the primary subject described in the sentence, carrying key semantic information. Through the ``[ADJ]-CCOMP-[NOUN]'' correlation, the emotional word ``exciting'' directly modifies ``team''. Compared to previous work that did not consider such correlations, the model is more likely to identify key semantic words that are directly associated with emotional words through these types of syntax-meaningful relationships. Therefore, focusing on the variability and relevance of emotional words can promote the correct recognition of emotions and the detection of important semantics.

Therefore, we propose a dynamical Emotion-Semantic Correlation Model (ESCM) for empathetic dialogue generation. 
ESCM dynamically constructs emotion-semantic vectors through the interaction of context and emotions. 
By encoding emotion-semantic vectors, the model dynamically adjusts emotions and semantics in the context to capture the variability of emotional words. 
To reflect the correlations between emotions and semantics clearly, we introduce a dependency tree. 
Based on the dynamic emotion-semantic representation and the dependency tree, ESCM proposes a dynamic correlation graph convolutional network to guide the model to capture the correlations between emotions and semantics clearly. 
By learning dynamic emotion-semantic representations and their correlations, ESCM accurately understands the emotions of the dialogue and captures important semantics to generate more empathetic responses.

We conduct experiments on the EMPATHETIC-DIALOGUES dataset~\cite{rashkin2018towards}. 
The results show that the ESCM model accurately understands the dialogue and generates grammatically fluent and informative empathetic responses. 
Furthermore, we extract and statistically analyze the common correlation structures in dialogues from the Empathetic-Dialogue dataset. 
The results indicate that the correlations between emotion and semantics are frequently and extensively utilized in expressing emotions during conversations. Additionally, the results of our analysis of correlation structures are consistent with linguistic conclusions ~\cite{foolen2012moving}.

To sum up, our contributions are as follows:

\begin{itemize}
\item 
We introduce the expressive characteristics of emotions in linguistics, including the variability of emotions and the correlations between emotions and semantics, to enhance the understanding of the meaning in conversations.
\item 
We propose the ESCM model, which constructs dynamic emotion-semantic vectors to adjust the dynamics of emotions, and leverages a dependency tree-based dynamic correlation graph convolutional network to learn correlations, in order to generate empathetic responses.
\item Experiments on the EMPATHETIC-DIALOGUE dataset demonstrate the effectiveness of ESCM. Furthermore, additional statistical and analytical experiments show that the correlations in dialogue are consistent with psychological research.
\end{itemize}

\section{Related Work}
Empathetic response generation refers to empathetically responding by perceiving emotional feelings in the language of the speaker~\cite{rashkin2018towards}. 

Early approaches explore the overall emotions of the conversation. 
~\citet{rashkin2018towards} introduce emotion representation generated by a pre-trained emotion classifier to learn and express specific types of emotions in the conversation. 
However, emotions expressed in responses are often diverse rather than specific ~\cite{lin2019moel}.
Therefore, ~\citet{lin2019moel} utilize multiple professional emotion listeners to express various appropriate emotions.
~\citet{majumder2020mime} group multiple conversation emotions by polarity and simulate the speaker's emotions to generate empathetic responses.

These methods focus on the overall emotions of the conversation and ignore nuanced emotions~\cite{li2019empdg}. 
To capture nuanced emotions, ~\citet{li2019empdg} extract emotional words through the NRC Emotion Lexicons~\cite{mohammad2013crowdsourcing} and integrate them into the model.
~\citet{gao2021improving} and ~\citet{Kim2021empathy} introduce emotional cause detection models to capture emotional words and perceive nuanced emotions.
~\citet{li-etal-2022-kemp} enhance emotional representation in the context with additional knowledge, which helps detect emotional words. 
~\citealt{CEM2021} use commonsense reasoning knowledge to infer nuanced emotions in the conversation. 
~\citealt{kim2022emp} employ pre-trained models to detect word-level emotion and keywords to detect the nuanced emotion in dialogues.
These methods detect emotional words with nuanced emotions in the language of the conversation and use static vectors such as word embeddings or VAD to represent emotional words.

Overall, early approaches ignore emotional words with nuanced emotions. 
Recent methods ignore two major characteristics of emotional words in linguistic expression: variability and correlation. 
Unlike these methods, we consider the two characteristics of emotional words and propose a dynamic emotion-semantic correlation model to better understand the conversation.
\begin{figure*}
\centering
\includegraphics[width=150mm]{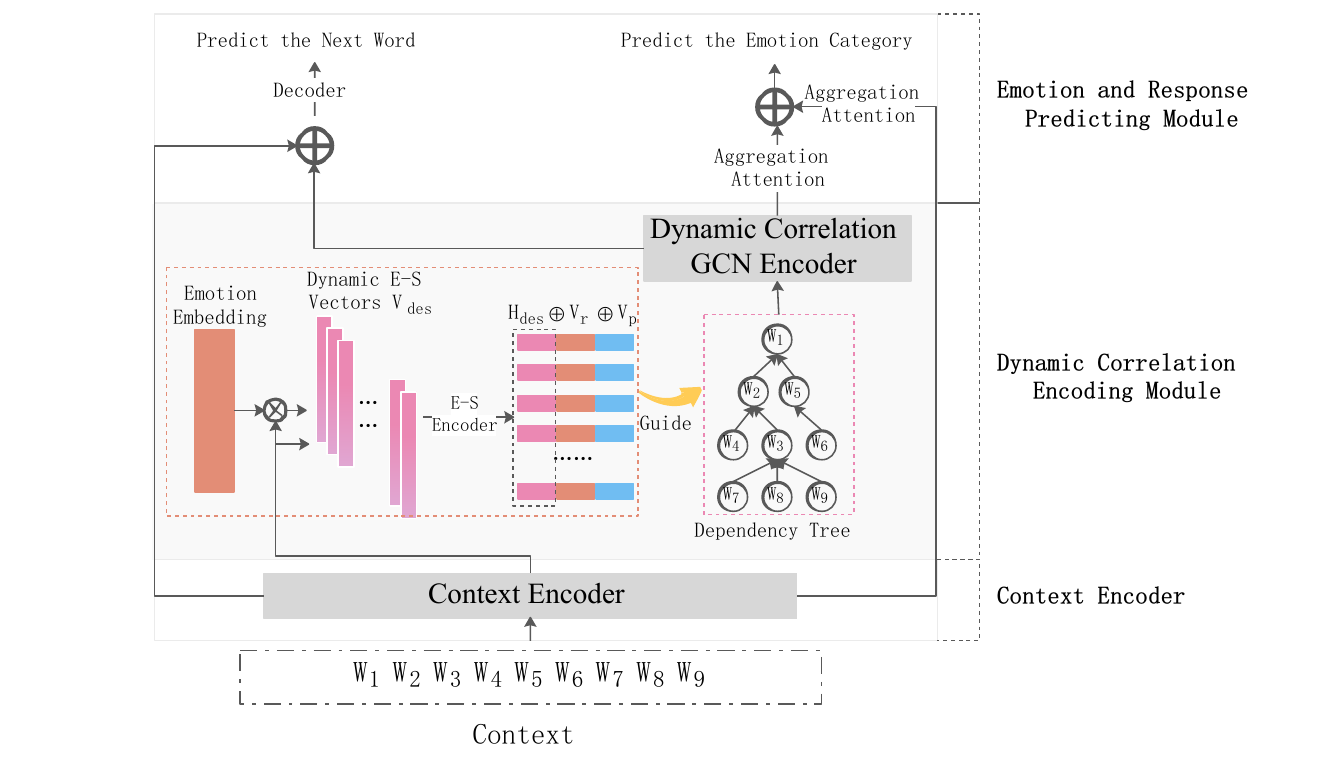}
\caption{\label{fig 2}
An overview of ESCM. ESCM consists of three main key modules: 
(1) a context encoder (\textbf{Section \ref{Context Encoder}}), which encodes the semantic of context. 
(2) a dynamic correlation encoding module (\textbf{Section \ref{Dynamic Correlation Encoding Module}}), which learns the correlations between emotions and semantics. 
(3) emotion and response predicting module (\textbf{Section \ref{Emotion and Response Predicting}}), which predicts dialog emotion categories and generates empathetic responses.
}
\end{figure*}
\section{Method}
\subsection{Task Formulation}
Given a dialogue context $D$ = [$U_1, U_2, ..., U_{M}$] of two interlocutors, our model needs to accurately perceive the emotions and semantics in the dialogue context and generate empathetic responses $Y$ = [$y_1$, $y_2$, ...,$y_j$, $y_N$]. Here, $U_i$ = [$w^i_1, w^i_2, ..., w^i_{m_i}$] represents the i-th utterance containing $m_i$ words. $Y$ is a response containing N words.

\subsection{Overview}
We propose ESCM, which takes into account the dynamic emotions and semantics in the dialogue and their correlations.
The proposed model is a transformer-based model with encoder-decoder architecture. 
To accurately perceive the content of the dialogue, we mainly reconstruct the encoder. 
As shown in Figure \ref{fig 2}, ESCM mainly consists of three parts: (1) a context encoder (\textbf{Section \ref{Context Encoder}}), which is a standard encoder structure and is used to understand the semantics of the dialogue; (2) a dynamic correlation encoding module (\textbf{Section \ref{Dynamic Correlation Encoding Module}}), which includes the construction of dynamic correlation vectors and the encoding of a dynamic correlation graph convolutional network. It learns the correlations between emotions and semantics; (3) emotion and response predicting module (\textbf{Section \ref{Emotion and Response Predicting}}), which completes the functions of emotion prediction and response generation.

\subsection{Context Encoder}
\label{Context Encoder}
As with previous methods~\cite{li2019empdg, li-etal-2022-kemp, CEM2021}, we concatenate the utterances of the dialogue context and prepend $[CLS]$ as the whole sequence token to form the context input $C$ =  = $[CLS] \oplus U_1 \oplus U_2 \oplus ... \oplus U_{M}$. Here, $\oplus$ denotes the concatenation symbol.
To input the context $C$ into the model, we convert $C$ into context word embeddings $E_c$.
Then, we sum up the word embeddings $E_c$, position embeddings, and state embeddings to form the semantic embeddings $\widetilde{E}_c$.
The state embeddings are used to distinguish between speaker or responder types and are randomly initialized.
To understand the semantics of the dialogue, we feed the semantic embeddings $\widetilde{E}_c$ into the context encoder $Enc_{ctx}$ to obtain the context semantic representation $H_{k}$:
\begin{gather}
H_{ctx} = Enc_{ctx}(\widetilde{E}_c)
\label{equation 1}
\end{gather}
where $H_{ctx} \in R^{L \times d}$, L is the length of the context sequence, and d represents the hidden size of the encoder.
\subsection{Dynamic Correlation Encoding Module}
\label{Dynamic Correlation Encoding Module}
Dynamic correlation encoding consists of two sub-modules:
(1) Dynamic emotion-semantic vectors. It brings the model the ability to flexibly adjust emotions and semantics, making the representation of context more reasonable.
(2) Dynamic correlation graph convolutional network. This module uses emotions, semantics, part-of-speech, and dependency types to guide the model to discover and aggregate words with strong correlations, in order to more accurately understand the emotions and semantics of the conversation.

\textbf{Dynamic Emotion-Semantic Vectors}.
Context has a significant impact on words, and understanding words without context may lead to errors. For example, without context, the word ``well'' is generally considered to have a positive emotion, but in sentence 1 (shown in Figure \ref{fig 1}), it functions as an interjection to enhance the tone. This usage does not indicate a positive meaning. Therefore, it is necessary to dynamically adjust emotions and semantics to adapt to the context.

Regarding semantics, we utilize weighted adjustments of context word embeddings with semantics.
\begin{gather}
    E_{ds} = w_{s} E_c + b_s
\end{gather}
where $w_s$ and $b_s$ are trainable parameters, and $E_{ds} \in R^{L \times d_s}$. $d_s$ is the hidden size of the dynamic semantic vector $E_{ds}$.

Regarding emotions, we interact context word embeddings $E_c$ with emotion embeddings $E_e$ to obtain dynamic emotion vectors $E_{de}$. Emotion embeddings refer to emotion categories represented in word form, which are converted into vector embeddings.
\begin{gather}
    E_{dot} = (w_c E_c + b_c) \cdot (w_e E_e + b_e)^T \\
    E_{de} = w_{ce} E_{dot} + b_{ce}
\end{gather}
where $w_c, b_c, w_e, b_e, w_{ce}, b_{ce}$ are trainable parameters. $E_{dot}, E_{de} \in R^{L \times d_{e}}$, and $d_{e}$ is the number of emotional categories.

We then feed the combined emotion and semantic vectors into the encoder to learn emotion-semantic representations. 
In this way, the model can take into account both emotion and semantics simultaneously during training to more comprehensively understand words in context.
\begin{gather}
    V_{des} = E_{de} \oplus E_{ds} \\
    H_{des} = Enc_{des}(V_{des})
    \label{equation 6}
\end{gather}
where $V_{des}$, $Enc_{des}$, and $H_{des}$ refer to the dynamic vectors, encoder, and representations for emotion-semantics, respectively.
And $V_{des}, H_{des} \in R^{L \times (d_s + d_{e})}$.

\textbf{Dynamic Correlation Graph Convolutional Network}.
The next problem is how to focus on and learn grammatical correlations. 
Dependency trees clearly reflect the grammatical dependency relationships between related words~\cite{kuncoro2016recurrent,kiperwasser2016simple,chen2014fast,dozat2016deep}.
Therefore, we use dependency trees to reflect the correlations between words. 
However, words with correlations in the dependency tree are not always important for understanding emotions and semantics.
For example, ``exciting'' and ``team'' are more important, while ``exciting'' and ``is'' are relatively unimportant.
To distinguish these correlations, we consider multiple aspects of correlation guidance, including dynamic emotion-semantic representation, part-of-speech of related words, and the dependency types between them.

We list two examples to illustrate the validity of the above correlation guidance.
In sentence 2 (shown in Figure \ref{fig 1}), the part-of-speech (ADJ and NOUN)\footnote{ADJ: adjective, NOUN: noun} and the dependency type (CCOMP)\footnote{CCOMP: clausal complement} indicate that ``exciting'' is closely related to the key semantic ``team''. Conversely, the part-of-speech (ADJ and AUX)\footnote{AUX: auxiliary} and dependency type (COP)\footnote{COP: copula} indicate that the correlation between ``exciting'' and ``is'' is trivial and unimportant.

Therefore, we consider the above multiple aspects and concatenate the emotion-semantic representation, part-of-speech, and dependency type to form the guiding vector $V_{qk}$.
\begin{gather}
        V_{qk} = H_{des} \oplus V_p \oplus V_r 
        \label{equation 7}
\end{gather}
where $V_p$, $V_r$ respectively denote part-of-speech embeddings , dependency type embeddings, which are randomly initialized. And $V_p, V_r \in R^{L \times L \times (d_{pr})}$, $V_{qk} \in R^{L \times L \times (d_s + d_e + 2d_{pr})}$. $d_{pr}$ is the embedding size of $V_p$ or $V_r$.

By assigning probabilities to each correlated neighbor, we aggregate the neighboring nodes that are associated in the dependency tree.
Subsequently, we obtain the correlation representation $H_{cor}$.
\begin{gather}
    p_{i, j} = \frac{a_{i, j} \cdot exp(V_{qk}[i] \cdot V_{qk}[j])}{\sum^{L}_{j=1}{a_{i, j} \cdot exp(V_{qk}[i] \cdot V_{qk}[j])}} \\
    H_{cor} = ReLU(\sum^{L}_{j=1}{p_{i, j}(W_v V_{des}[j] + b_v)})
    \label{equation 9}
\end{gather}
where $p_{i, j} \in R^{L \times L \times 1}$, $H_{cor} \in R^{L \times (d_s + d_e)}$. $a_{i, j}$ is the value of the adjacency matrix about the dependency tree. When node i and node j have a direct relationship in the dependency tree, $a_{i, j}$ is 1, otherwise it is 0. $V_{qk}[i]$ and $V_{qk}[j]$ represent the guiding vectors of node i and node j, respectively. $W_v$ and $b_v$ are trainable parameters, and $ReLU$ is the ReLU activation function. 

\subsection{Emotion and Response Predicting}
\label{Emotion and Response Predicting}
Based on the context semantics and the correlations between emotions and semantics, we predict the emotions of the conversation and generate empathetic responses.

\textbf{Emotion Predicting}.
To understand the context semantics and capture important correlations, we use two aggregation networks with the same structure but different parameters to process the context semantic representation $H_{ctx}$ (Eq. \ref{equation 1}) and correlation representation $H_{cor}$ (Eq. \ref{equation 9}). We take the processing of the context representation as an example.

We first calculate the weights of the words in the context semantic representation $H_{ctx}$ and sum them up according to their weights to obtain the hidden layer representation $H^2$.
\begin{gather}
    H^{1}_{a} = Tanh(w^1_{a} H_{ctx} + b^1_{a}) \\
    P_{s} = Softmax(w^{1}_{s} H^{1}_{a} + b^{1}_{s}) \\
    H^2 = \sum^L_{j=1} P_{s}[j] \cdot H_{ctx}[j]
\end{gather}
where $H^{1}_{a} \in R^{L \times d}$, $P_{s} \in R^{L \times 1}$, $H^2 \in R^d$. $w^1_a, b^1_a, w^1_s, b^1_s$ are learnable parameters, and $Tanh$ is the tanh activation function.

Then we feed the hidden layer representation into a non-linear layer to learn and generate context semantic emotion probabilities $P^e_{ctx}$.
\begin{gather}
    H^2_{a} = Tanh(w^2_{a} H^2 + b^2_{a}) \\
    P^e_{ctx} = Softmax(w^{2}_{s} H^{2}_{a} + b^{2}_{s})
\end{gather}
where $H^2_a \in R^d$, $P^e_{ctx} \in R^d_{e}$, $w^2_a, b^2_a, w^2_s, b^2_s$ are learnable parameters.

Similarly, we use the same structure to construct an aggregation attention network about the correlations and obtain emotion probabilities $P^e_{cor} \in R^d_{e}$. We add the two types of emotion probabilities together as the overall emotion probability $P_e \in R^d_{e}$.
\begin{gather}
    P_e = P^e_{ctx} + P^e_{cor}
\end{gather}

To ensure that important information in semantics and correlations is not affected by each other, we set loss functions for them separately.
We employ log-likelihood loss to optimize the parameters during the training phase based on the emotion category and the ground truth label.
\begin{gather}
    \mathcal{L}^e_{ctx} = - log(P^e_{ctx}(e^*)) \\
    \mathcal{L}^e_{cor} = - log(P^e_{cor}(e^*))
\end{gather}

\textbf{Response Predicting}.
Similarly, our decoder generates responses based on the context semantics and the correlations between emotion and semantics. This module takes the context semantic representation $H_{ctx}$ and the correlation representation $H_{cor}$ as input and predicts the next word at each time step t. Similar to ~\cite{li-etal-2022-kemp}, we use a point generator network to capture key vocabulary in the context and correlations.
\begin{gather}
    H = H_{ctx} \oplus H_{cor} \\
    P(y_t|y<t, C) = Dec(E_{y<t}, H)
\end{gather}
where $H \in R^{L \times (d + ds + de)}$, $P(y_t|y<t, C) \in R^{V}$. $V$ is the length of the vocabulary. $Dec$ represents a decoder with a pointer network.

Subsequently, we use cross-entropy as generation loss.
\begin{gather}
    \mathcal{L}_{gen}(y_t) = - \sum^T_{t=1}log(P(y_t|y<t, C))
\end{gather}

\textbf{Total Loss}.
Finally, we add the loss $\mathcal{L}_{gen}(y_t)$ and two emotion losses $\mathcal{L}^e_{ctx}$ and $\mathcal{L}^e_{cor}$ together to obtain the total loss $\mathcal{L}$. 
We optimize the training parameters in the model using the total loss.
\begin{gather}
    \mathcal{L} = \mathcal{L}_{gen}(y_t) + \mathcal{L}^e_{ctx} + \mathcal{L}^e_{cor}
\end{gather}
\section{Experiments}

\subsection{Baselines}
We compare recent state-of-the-art baselines with our model.

\textbf{Transformer}~\cite{vaswani2017attention} is a vanilla Seq2Seq model, including both encoder and decoder;

\textbf{EmoPrend-1}~\cite{rashkin2018towards} is a Transformer-based model that enhances empathy by incorporating emotion labels from a pre-trained emotion classifier;
 
\textbf{MoEL}~\cite{lin2019moel} is also a Transformer-based model that 
softly combines various emotions with multiple decoders to generate empathetic responses;

\textbf{MIME}~\cite{majumder2020mime} is a Transformer-based model, which 
consider polarity-based emotion clusters and emotional mimicry to generate appropriate responses;

\textbf{EmpDG}~\cite{li2019empdg} 
emphasizes the importance of user feedback and multi-resolution emotions. It uses a generative adversarial network to train the model and generate empathetic responses;

\textbf{KEMP}~\cite{li-etal-2022-kemp} 
employs ConceptNet as extra knowledge to enrich the representation of implicit emotions and captures these emotions to generate appropriate responses;

\textbf{CEM}~\cite{CEM2021} 
 takes into account both the emotional and cognitive aspects of empathy. By incorporating reasoning knowledge, it enhances the ability to perceive and express emotions.

\subsection{Implementation Details}
We conduct experiments on the EMPATHETIC-DIALOGUES~\cite{rashkin2018towards} dataset. 
In the dataset, the number of emotions is $d_e$=32.
In the model, we use Glove~\cite{pennington2014glove} as the initialization vector for word embedding, with a dimension of d=300.
We set the dimension of the dynamic emotion vector to $d_s$=10.
At the same time, the dimensions of the part-of-speech embedding and dependency type embedding are both set to $d_{pr}$=50.
We use Biaffine Parser~\cite{dozat2016deep} to obtain dependency relationships. 
For the multi-head attention networks in our model, we use a 1-layer network and set the number of heads to 2.
Subsequently, we set the batch size to 16 and use the Adam optimizer~\cite{kingma2014adam} to optimize the parameters.
After training for 13500 rounds on an NVIDIA Tesla T4 GPU, the model converged.

\subsection{Evaluation Metrics}
As with previous work, we employ both automatic and manual metrics to evaluate the performance of the model.

\textbf{Automatic Evaluation Metrics}. Following ~\cite{li-etal-2022-kemp, CEM2021}, we use the following automatic metrics in our experiments: Perplexity (PPL), Accuracy (Acc), Dist-1, and Dist-2.
PPL measures language fluency, which is of higher quality when the score is lower. Acc assesses the accuracy of emotion perception. Dist-1 and Dist-2~\cite{li2015diversity} measure response diversity at single and double granularity, respectively.

\begin{table}
\centering
\begin{tabular}{ccccc}
\hline
\textbf{Models} & \textbf{Acc} & \textbf{PPL} & \textbf{Dist-1} & \textbf{Dist-2} \\
\hline
Transformer & - & 37.73 & 0.47 & 2.04 \\ 
EmoPrend-1 & 33.28 & 38.30 & 0.46 & 2.08 \\ 
MoEL & 32.00 & 38.04 & 0.44 & 2.10 \\ 
MIME & 34.24 & 37.09 & 0.47 & 1.91 \\ 
EmpDG & 34.31 & 37.29 & 0.46 & 2.02 \\ 
KEMP & 39.31 & 36.89 & 0.55 & 2.29 \\ 
CEM & 39.11 & 36.11 & 0.66 & 2.99 \\ 
\hline
ESCM & \textbf{41.19} & \textbf{34.82} & \textbf{1.19} & \textbf{4.11} \\ 
\hline
\end{tabular}
\caption{\label{table 1}
The automatic evaluation results.
}
\end{table}
\textbf{Human Evaluation Metrics}. 
Previous work scores models' responses on a scale of 1 to 5 to assess their quality~\cite{li2019empdg,li-etal-2022-kemp}. This type of assessment is prone to inconsistent results due to differences in individual criteria ~\cite{CEM2021}.
Therefore, we adopt the A/B test strategy ~\cite{lin2019moel, majumder2020mime}. 
Given two responses generated by the models for the same conversation, three professional crowdsourcers are required to assign ESCM a score of 1 on \emph{Win} when the response generated by ESCM is better than the compared model. 
Correspondingly, when ESCM is better than or equal to the compared model, the crowdsourcers will add points for ESCM or \emph{Tie}.
Furthermore, three aspects are considered for evaluating models: empathy, relevance, and fluency.
Empathy evaluates whether the responses show the right types of emotions; 
Relevance measures whether the reply is consistent with the theme and semantics of the context;
Fluency assesses the response's readability and grammatical accuracy.
\begin{table}
\centering
\begin{tabular}{ccccc}
\hline
\textbf{Comparisons} & \textbf{Aspects} & \textbf{Win} & \textbf{Lose} & \textbf{$\kappa$} \\
\hline
\multirow{3}{*}{\centering \shortstack{ESCM \\ vs. EmpDG}} & Emp. & \textbf{45.4} & 24.0 & 0.48\\
& Rel. & \textbf{52.8} & 16.3 & 0.43 \\
& Flu. & \textbf{50.1} & 5.9  & 0.45 \\
\hline
\multirow{3}{*}{\centering \shortstack{ESCM \\ vs. KEMP}} & Emp. & \textbf{44.0} & 20.0 & 0.57\\
& Rel. & \textbf{53.3} & 21.0 & 0.46 \\
& Flu. & \textbf{35.4} & 13.4 & 0.41 \\
\hline
\multirow{3}{*}{\centering \shortstack{ESCM \\ vs. CEM}} & Emp. & \textbf{37.3} & 19.8 & 0.58 \\
& Rel. & \textbf{48.9} & 21.5 & 0.41 \\
& Flu. & \textbf{33.8} & 11.6 & 0.47 \\
\hline
\end{tabular}
\caption{\label{table human}
Results of human evaluation. 
For a more intuitive display, we remove the result of the Tie and only show Win and Lose.
Where $\kappa$ is the inter-labeler agreement measured by Fleiss's kappa~\cite{fleiss1973equivalence}, and 0.4 $\textless$ $\kappa$ $\leq$ 0.6 indicates moderate agreement.
}
\end{table}
\section{Results and Analysis}
\subsection{Main Results}
\begin{table}
\centering
\begin{tabular}{cccccc}
\hline
\textbf{Models} 
& \textbf{Acc} & \textbf{PPL} & \textbf{Dist-1} & \textbf{Dist-2}\\
\hline
ESCM 
& \textbf{41.19} & 34.82 & \textbf{1.19} & \textbf{4.11}\\
\hline
w/o DESV
& 39.21 & 34.10 & 1.07 & 3.52\\
\hline
w/o DCGCN
& 39.05 & \textbf{33.52} & 1.08 & 3.69\\

w/o $V_r$
& 40.0 & 34.02 & 1.06 & 3.68\\

w/o $V_p$
& 39.41 & 34.45 & 0.99 & 3.33\\

w/o $V_{des}$
& 40.42 & 34.48 & 1.08 & 3.60\\
\hline
\end{tabular}
\caption{\label{table 5}
Results of the ablation experiments.
}
\end{table}

\textbf{Automatic Evaluation Results}. Table \ref{table 1} shows the main results of the automatic evaluation for all models.
We find that early models (EmoPrend-1, MoEL, and MIME) are not as effective as models that focus on subtle emotions (EmpDG, KEMP, CEM). 
Additionally, we find that ESCM outperforms the baselines in all metrics. 
In terms of diversity, ESCM significantly outperforms the baselines. 
This suggests that focusing on the correlations between emotions and semantics helps the model capture key semantics and express informative responses. ESCM also outperforms the baselines in terms of emotion accuracy, indicating the effectiveness of the dynamic emotion-semantic vectors. 
Furthermore, ESCM achieves the best fluency, which indicates that the model combines emotions and semantics to express more natural language.

\textbf{Human Evaluation Results}. 
As shown in Table \ref{table human}, ESCM outperforms the three strongest baselines in terms of empathy, relevance, and fluency. 
The superiority in empathy indicates that the model accurately understands and expresses emotions. 
The significant improvement in relevance suggests that the model captures and expresses key semantics. 
The superiority in fluency indicates that the model expresses more fluent responses by better understanding the context.

\subsection{Ablation Studies}

To verify the effectiveness of each component, the following experiments are conducted:

(1) \textbf{w/o DESV}: Dynamic emotion-semantic representations $H_{des}$ (in Eq. \ref{equation 6}) are replaced by context semantic representations $H_{ctx}$ (in Eq. \ref{equation 1});

(2) \textbf{w/o DCGCN}: No dynamic correlation graph convolutional network (in Eqs. \ref{equation 7} - \ref{equation 9});

(3) \textbf{w/o $V_r$/$V_p$/$V_{des}$}: Without the guidance of vectors $V_r$/$V_p$/$V_{des}$ (in Eq. \ref{equation 7}) in the dynamic correlation graph convolutional network.

The results of the ablation experiments are shown in Table \ref{table 5}. 
To verify the effectiveness of the dynamic emotion-semantic vectors, we remove $DESV$. 
The results show a significant decrease in emotional accuracy and diversity.
The drop in emotion accuracy suggests that dynamic emotion-semantic vectors play an important role in capturing emotions.
The changes in diversity metrics demonstrate the crucial role of dynamic adjustment of emotions and semantics in precise understanding and informative expression in conversations.

We remove $DCGCN$ and its various guiding vectors to verify the effectiveness of the dynamic correlation graph convolutional network. 
After removing $DCGCN$, we find a significant decrease in emotional accuracy and diversity.
This indicates that correlations have a significant impact on the perception of emotions and semantics, and they play an important role in expressing informative responses. 
We further explore the role of various guiding vectors. 
When part-of-speech $V_p$ or dependency types $V_r$ are removed, the emotional accuracy decreases significantly.
This indicates that ESCM can aggregate effective information related to emotions based on part-of-speech or dependency types. 
After removing part-of-speech, dependency types, and emotion-semantic vectors respectively, the diversity decreases significantly. 
This indicates that these features affect the aggregation of semantic information. 

Furthermore, we find that removing any module improves the fluency of responses but decreases diversity. 
This is mainly due to the fact that the ablation models express responses with more fluent yet less information, such as trivial sentences.
\subsection{Correlation Analysis}
To further explore the correlations, we conduct a statistical analysis of the correlations in the EMPATHETIC-DIALOGUES dataset (see appendix \ref{sec:appendix} for details). 
We extract 1138 correlations from the dialogue dataset, which are used a total of 151242 times in the dialogue. 
This indicates that the correlation structure is frequently and repeatedly used in the dialogue.
In addition, we find that emotions are mainly expressed through three parts of speech: adjectives, nouns, and verbs. 
This is consistent with linguistic research ~\cite{foolen2012moving} on the parts of speech of emotional words. 
At the same time, we also find that the ``preposition + noun'' structure is frequently used, which is also consistent with linguistic research ~\cite{foolen2012moving}. 
In each emotion type of empathetic dialogue, the frequently used correlation structures are similar, but the frequency of use may differ.

\subsection{Time and Resource Consumption}
To further demonstrate the efficacy of the model, we conduct analytical experiments on time and resource consumption.

\textbf{Time Consumption}. As shown in Table \ref{table time}, we calculate the time for our model and baselines to converge during training. The results show that our model does not have significantly higher average per-iteration time compared to the baselines. This is mainly because: although the intermediate variable $V_{qk}$ intuitively has higher dimensionality, in practice the dialogue context length L is short, so the impact on time is not substantial. Meanwhile, since our model can better understand the dialogue, it requires fewer iterations. Therefore, the total training time is actually significantly less than the baselines.

\textbf{Resource Consumption}. As shown in Table \ref{table gpu}, we compute the GPU memory consumption of the models. Due to the higher dimensionality of the intermediate variable $d_{pr}$ in our model, our model requires slightly more resources than the baselines (see last row). However, even when reducing the GPU consumption of our model to be comparable to the baselines, our model still significantly outperforms the baselines overall (see penultimate row).
\begin{table}
\centering
\begin{tabular}{cccc}
\hline
\textbf{Models} & \textbf{T1 (s)} & \textbf{T2 (s)} & \textbf{Num}\\
\hline
KEMP & 0.17 & 4378.11 & 26,000 \\ 
CEM & 0.22 &  4438.68 & 20,000 \\ 
\hline
ESCM & 0.20 & 2733.94 & 13,500 \\ 
\hline
\end{tabular}
\caption{\label{table time}
Results of time consumption. T1 represents the average per-iteration time, T2 stands for the convergence time, and Num is the number of iterations needed for convergence.
}
\end{table}

\begin{table}
\centering
\begin{tabular}{p{2.1em}p{1.5em}p{1.5em}p{2.6em}p{2.6em}p{2.6em}}
\hline
\textbf{Models} & \textbf{PPL} & \textbf{Acc} & \textbf{Dist-1} & \textbf{Dist-2} & \textbf{Size(G)}\\
\hline
KEMP & 39.31 & 36.89 & 0.55 & 2.29 & 6.02 \\
CEM & 39.11 & 36.11 & 0.66 & 2.99 & 5.57 \\
V1 & 33.74 & 40.21 & 0.98 & 3.17 & 4.64 \\
V2 & 34.82 & 41.63 & 1.07 & 3.52 & 6.00 \\
V3 & 34.82 & 41.19 & 1.19 & 4.11 & 8.10 \\
\hline
\end{tabular}
\caption{\label{table gpu}
Results under restricted resource consumption. `Size' denotes the GPU memory required to train the model. We list three ESCM models with different $d_{pr}$ values, where V1, V2, and V3 represent models with $d_{pr}=10$, $d_{pr}=20$, and $d_{pr}=50$, respectively.
}
\end{table}

\subsection{Case Study}
\begin{table}
\centering
\begin{tabular}{p{1.3cm}|p{5cm}}
\hline
\textbf{Emotion} & \textbf{Annoyed}\\
\textbf{Conext} & I \textcolor{c7}{hate} when my little brother throws food on the table just to \textcolor{c8}{annoy} me\\
\textbf{Gold} & That is a \textcolor{c7}{terrible} \textcolor{c7}{waste} of food, how old is your little brother? \\
\hline
\textbf{CEM} & I would be \textcolor{c9}{mad} too\\
\textbf{KEMP} & I am \textcolor{c7}{sorry} to hear that. I hope you can get it back soon.\\
\hdashline
\textbf{ESCM} & \textbf{That is really \textcolor{c9}{annoying}. I \textcolor{c7}{hate} when that happens.}\\
\hline
\textbf{Emotion} & \textbf{Proud}\\
\textbf{Conext} & I am constantly \textcolor{c7}{amazed} by my sister's art skills. I am not sure where she got them ... my brother and I can not draw to save our lives. \\
\textbf{Gold} & What does she draw?\\
\hline
\textbf{CEM} & That is \textcolor{c7}{awesome}! I bet she's a \textcolor{c9}{great} job!\\
\textbf{KEMP} & I am sure you will do \textcolor{c7}{great}!\\
\hdashline
\textbf{ESCM} & \textbf{That is really \textcolor{c7}{cool}. I bet you are very \textcolor{c9}{proud} of her.}\\
\hline
\end{tabular}
\caption{\label{table 7}
Case Study of ESCM and benchmarks. Words with Marked colors in the sentence are rich in emotion.
}
\end{table}
As shown in Table \ref{table 5}, we select two strongest baselines and compare them with ESCM through sample analysis. 
In the first case, The speaker is annoyed due to the fact that ``the younger brother threw food onto the table''.
The baselines do not accurately understand the emotional expression ``annoyed'' and the event it described. 
However, ESCM understands and expresses ``annoying'' correctly, and gives a response indicating disgust towards the event. 
This indicates that ESCM is able to capture key semantics through correlations. 
In the second case, the speaker expresses the emotion of ``proud'' using the word ``amazed'' with a surprised emotion. 
The baselines do not understand the emotions and semantics involved, while ESCM accurately understands the emotions and expresses an empathetic response. This demonstrates the effectiveness of building dynamic emotion-semantics.

\section{Conclusion and Future Work}
This paper proposes ESCM, which introduces two characteristics of emotions in the linguistic expression process: variability and the correlations between emotions and semantics. The proposed model constructs a dynamic emotion-semantic vector to reflect variability and uses a dependency tree-based dynamic correlation graph convolutional network to learn correlations. Both automatic and manual metrics demonstrate the effectiveness of the model.
Furthermore, we conduct statistical analysis experiments. The results show that correlations are frequently used in the dialogue. Additionally, we find that the correlation structures in the dialogue are consistent with linguistic research.

To further investigate the correlation between emotion and semantics, we will take into account pre-trained knowledge, multilinguality, personalization, and other factors in future work.

\section*{Limitations}
The limitations of our work are as follows: 
(1) Pre-trained models have become the mainstream nowadays. To further explore the impact of pre-trained models, we constructed a pre-trained ESCM model. Since the word coverage of the EMPATHETIC-DIALOGUES dataset in the vocabulary of pre-trained models is only 51.8\%, we only surpassed the baseline Emp-RFT~\cite{kim2022emp} on two metrics (Acc: 42.44 (42.08), Dist-2: 9.91 (4.48)). In the future, we will further explore the impact of pre-trained models on correlations.
(2) The correlations between emotions and semantics discussed in this paper are only applicable to English. However, different languages may have different types of correlations~\cite{foolen2012moving}. Therefore, we will investigate the correlations between emotions and semantics in multilingual contexts in the future.
(3) Intuitively, there are individual differences in the expression of emotions. Due to data limitations, we did not consider this personalized factor. We will involve more research on correlations and personalization in the future.

\section*{Ethical Considerations}
The potential ethical implications of our work are as follows: (1) Dataset: EMPATHETICDIALOGUES is an open-source, publicly available dataset for empathetic response generation. In the dataset, the original provider has filtered information about personal privacy and ethical implications ~\cite{rashkin2018towards}. (2) Models: Our baselines are also open source, and they have no permission issues. Since our model is trained on a healthy dataset, it does not generate discriminatory, abusive, or biased responses to users.

\section*{Acknowledgments}
We would like to express our sincere gratitude to the reviewers for their diligent evaluation and constructive feedback, which helped improve the quality of this paper.
In addition, we appreciate the insightful discussions and comments from the authors, which stimulated valuable thinking around this work. Their diverse perspectives and experience shared through feedback have contributed immensely to the development of this research.
This work was supported by National Natural Science Foundation of China (No.61976054).

\bibliography{custom}
\bibliographystyle{acl_natbib}

\appendix
\section{Appendix}
\label{sec:appendix}
\begin{figure}
\centering
\includegraphics[width=82mm]{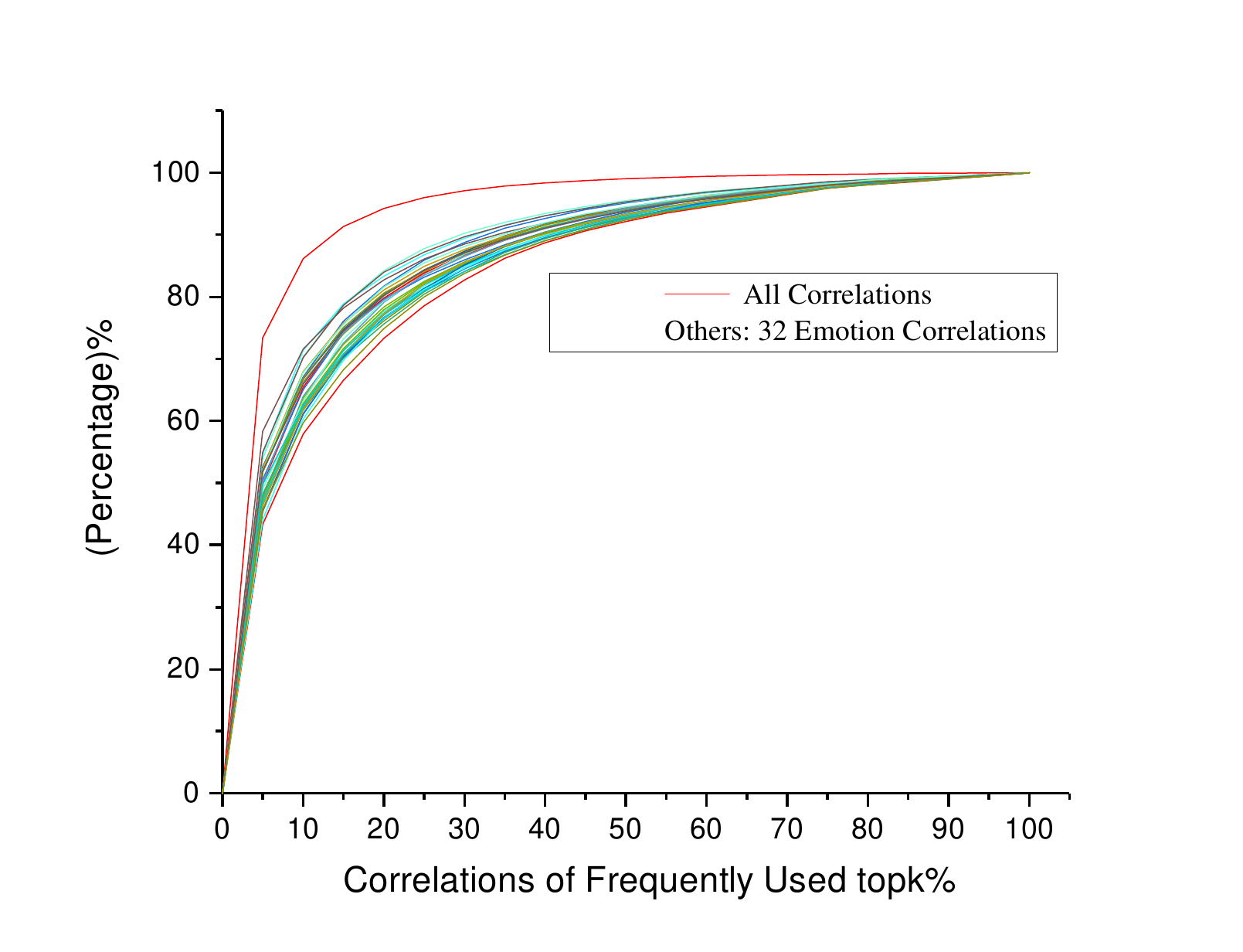}
\caption{\label{fig correlation_percentage}
The percentage of frequently used correlations to the total number of correlations.
}
\end{figure}

\begin{table}
\centering
\begin{tabular}{cc}
\hline
\textbf{Correlations/Percentage} & \textbf{Examples}\\
\hline
ROOT-root-ADJ-b(7.22) & \textcolor{c7}{joyful}, \textcolor{c7}{scary} \\
NOUN-amod-ADJ-b(6.51) & time \textcolor{c7}{good}\\ 
VERB-dep-ADJ-b(4.66) & felt \textcolor{c7}{proud}\\ 
ADP-pobj-NOUN-b(3.89) & into \textcolor{c7}{accident}\\
ADJ-nsubj-PRON-f(3.4) & \textcolor{c7}{nice} it\\
\hline
\end{tabular}
\caption{\label{table 5}
Top 5 correlations that are frequently used in conversations.
}
\end{table}

\textbf{Frequency Phenomenon}. 
As shown in Figure \ref{fig correlation_percentage}, we list the correlation statistical results.
(1) As shown in ``All Correlation'' in Figure \ref{fig correlation_percentage} (the highest red line), for the overall correlations in the dataset, the top 10\% of frequently used correlations account for more than 80\% of the total number of correlations. That is, $\frac{C_{f}}{C_{total}}$=86.17\%, where $C_f$ is the number of frequently used correlations in the top 10\%, and $C_{total}$ is the total number of correlations used in the dataset.

(2) As shown in ``Others: 32 Emotion Correlations'' in Figure \ref{fig correlation_percentage}, for each type of emotional empathetic dialogue, the top 20\% of frequently used correlations account for 80\% of the total number of correlations. That is, $\frac{C^e_{f}}{C^e_{total}}$ is approximately 80\%. For example, the number of empathetic dialogues expressing ``joyful'' emotion is $C^{joyful}_{total}$=6083, and the number of frequently used associations in the top 20\% is $C^{joyful}_{f}$=4945, so $C^{joyful}_{f} / C^{joyful}_{total}$ is approximately 81.13\%.

\begin{table*}[h!]
\centering
\begin{tabular}{ccc}
\hline
\textbf{Type} & \textbf{Top1} & \textbf{Top2}\\
\hline
surprised(8281)& ROOT-root-ADJ-b(6.48)& NOUN-amod-ADJ-b(5.72) \\
excited(6471)& NOUN-amod-ADJ-b(8.27)& ROOT-root-ADJ-b(7.87) \\
annoyed(4210)& ROOT-root-ADJ-b(7.01)& ROOT-root-VERB-b(5.11) \\
proud(5915)& ROOT-root-ADJ-b(11.78)& NOUN-amod-ADJ-b(5.93) \\
angry(4859)& ROOT-root-ADJ-b(10.91)& ADJ-nsubj-PRON-f(5.97) \\
sad(3945)& NOUN-amod-ADJ-b(7.25)& ROOT-root-ADJ-b(5.65) \\
grateful(6487)& ROOT-root-ADJ-b(8.89)& NOUN-amod-ADJ-b(6.98) \\
lonely(3083)& NOUN-amod-ADJ-b(11.39)& ADP-pobj-NOUN-b(6.94) \\
impressed(5483)& ROOT-root-ADJ-b(9.5)& NOUN-amod-ADJ-b(7.5) \\
afraid(5276)& ROOT-root-ADJ-b(6.99)& NOUN-amod-ADJ-b(4.78) \\
disgusted(4326)& ROOT-root-ADJ-b(7.19)& VERB-dep-ADJ-b(5.27) \\
confident(4203)& VERB-dep-ADJ-b(7.52)& ROOT-root-ADJ-b(7.47) \\
terrified(5436)& ROOT-root-ADJ-b(5.98)& NOUN-amod-ADJ-b(4.67) \\
hopeful(4635)& NOUN-amod-ADJ-b(10.68)& ROOT-root-ADJ-b(7.62) \\
anxious(4946)& ROOT-root-ADJ-b(8.07)& VERB-dep-ADJ-b(6.29) \\
disappointed(3861)& NOUN-amod-ADJ-b(7.15)& ROOT-root-ADJ-b(6.81) \\
joyful(6083)& ROOT-root-ADJ-b(8.83)& NOUN-amod-ADJ-b(5.85) \\
prepared(3617)& ROOT-root-ADJ-b(6.99)& ADP-pobj-NOUN-b(6.58) \\
guilty(4979)& VERB-dep-ADJ-b(12.07)& VERB-acomp-ADJ-b(8.44) \\
furious(4978)& ROOT-root-ADJ-b(9.62)& ADJ-nsubj-PRON-f(4.74) \\
nostalgic(4012)& NOUN-amod-ADJ-b(9.92)& ROOT-root-ADJ-b(4.99) \\
jealous(5210)& NOUN-amod-ADJ-b(7.98)& ROOT-root-ADJ-b(6.93) \\
anticipating(4502)& NOUN-amod-ADJ-b(8.93)& ROOT-root-ADJ-b(6.13) \\
embarrassed(3289)& ROOT-root-ADJ-b(5.17)& VERB-dep-ADJ-b(4.68) \\
content(6632)& ROOT-root-ADJ-b(8.05)& NOUN-amod-ADJ-b(7.86) \\
devastated(3905)& ADP-pobj-NOUN-b(6.17)& ROOT-root-ADJ-b(4.87) \\
sentimental(3123)& NOUN-amod-ADJ-b(8.65)& ROOT-root-ADJ-b(6.56) \\
caring(4041)& ROOT-root-ADJ-b(6.68)& NOUN-amod-ADJ-b(5.57) \\
trusting(4193)& NOUN-amod-ADJ-b(6.65)& ROOT-root-ADJ-b(6.39) \\
ashamed(3758)& VERB-dep-ADJ-b(9.05)& NOUN-amod-ADJ-b(4.68) \\
apprehensive(4267)& NOUN-amod-ADJ-b(8.39)& ROOT-root-ADJ-b(7.05) \\
faithful(3236)& NOUN-amod-ADJ-b(8.0)& ROOT-root-ADJ-b(5.87) \\
\hline
\end{tabular}
\caption{\label{table 6}
Top 2 frequently used correlations in conversations for each emotional dialogue. The numbers in the first column indicate the total number of correlations for the emotional dialogue, while the numbers in the other columns represent the percentage of times the correlations are used.
}
\end{table*}

\textbf{Part-of-Speech Phenomenon}. 
As shown in Table \ref{table 5}, we list examples of the most commonly used correlation structures. 
Taking ``NOUN-amod-ADJ-b'' as an example, ``NOUN-amod-ADJ'' represents a noun and an adjective linked together by the ``amod'' dependency type. 
``b'' refers to the emotion word as the second word, and ``f'' indicates that the emotion word is the first word. 
Specifically, ``ROOT'' refers to the root node of the dependency tree. 
In the frequently used correlations, emotions are mainly expressed through three parts of speech: adjectives, nouns, and verbs, which is consistent with linguistic research ~\cite{foolen2012moving} on emotion words.

\textbf{Correlation Structure Phenomenon}. 
As shown in Table \ref{table 5}, we list the most commonly used correlation structures for each type of emotion. We also find that the ``preposition + noun'' method is frequently used, which is consistent with linguistic research ~\cite{foolen2012moving}. 

As shown in Table \ref{table 6}, we list the most commonly used correlations in various emotional dialogues, with the numbers in parentheses indicating the probability of their usage. In each type of emotional empathetic dialogue, the frequently used correlation structures are similar, but the frequency of use may vary.

\end{document}